\documentclass[conference]{IEEEtran}
\usepackage{cite}
\usepackage{amsmath,amssymb,amsfonts}
\usepackage{graphicx}
\usepackage{textcomp}
\usepackage{xcolor}
\usepackage{booktabs}
\usepackage{multirow}
\usepackage{url}
\usepackage{algorithm}
\usepackage{algpseudocode}
\usepackage{amsmath}
\usepackage{amssymb}
\usepackage{booktabs}
\usepackage{hyperref}
\usepackage{subcaption}
\usepackage{tikz}
\usetikzlibrary{shapes,arrows,positioning,calc,fit,backgrounds}
\newcommand{\experimentfigure}[2][]{%
    \IfFileExists{#2}{\includegraphics[#1]{#2}}{%
        \fbox{\parbox[c][3.0cm][c]{0.94\linewidth}{\centering
        Experiment figure unavailable:\\[0.3em]\texttt{\detokenize{#2}}}}%
    }%
}

\def\BibTeX{{\rm B\kern-.05em{\sc i\kern-.025em b}\kern-.08em
    T\kern-.1667em\lower.7ex\hbox{E}\kern-.125emX}}

\begin{document}

\title{GAN-Diff : Coupling Pretrained WGAN-GP Features with Conditional Diffusion U-Nets}

\author{
    \IEEEauthorblockN{
        Saif Ahmed,
        Asadullah Hil Galib,
        S.M. Riaz Rahman Antu, 
        Ahmed Faizul Haque Dhrubo\textsuperscript{*},  \\ 
        Souvik Pramanik,
        Mohammad Abdul Qayum, 
        Mohsin Sajjad, 
        and Mohammad Ashrafuzzaman Khan
    }
    \IEEEauthorblockA{
        \textit{Department of Electrical and Computer Engineering}\\
        \textit{North South University},Dhaka, Bangladesh \\
        {\{saif.ahmed03, asadullah.galib01, riaz.antu, ahmed.dhrubo}, \\{souvik.pramanik, mohammad.qayum, mohsin.sajjad, mohammad.khan02\}}@northsouth.edu
    }
}

\maketitle

\begin{abstract}
Generative adversarial networks (GANs) can provide efficient image generation, while diffusion models offer high-quality image restoration but require iterative sampling. This paper presents a hybrid GAN-guided diffusion framework that uses a pretrained Wasserstein GAN with gradient penalty (WGAN-GP) as a feature prior for conditional diffusion-based image restoration. Intermediate features from the frozen WGAN-GP generator are incorporated into a diffusion U-Net through cross-attention and remain fixed during the DDIM sampling process. The framework is evaluated on two restoration tasks, Gaussian denoising and $2\times$\textit{super-resolution}, using CelebA face images. During development, several sources of instability were identified and addressed, including adversarial learning-rate imbalance, inappropriate diffusion initialization, excessive corruption, and insufficient parameter averaging. The resulting framework consistently improves the quality of both degraded and low-resolution images. In particular, it improves denoising performance by 4.40 dB in PSNR and super-resolution performance by 3.70 dB over their respective input baselines. These results demonstrate the potential of a frozen GAN feature prior to guide diffusion models toward stable and effective image restoration.
\end{abstract}


\section{Introduction}
Deep generative models have emerged as essential tools for image synthesis and recovery, and two types of such models, namely generative adversarial networks (GANs) and diffusion models, complement each other. In particular, GANs \cite{goodfellow2014gan} generate images using only one forward pass and thus provide efficient inference. Objective functions based on Wasserstein distance with gradient penalty, for example WGAN-GP \cite{gulrajani2017wgangp}, also provide additional benefits in terms of stable training. On the other hand, GANs may still exhibit mode collapse, oscillations during training, and instability due to the interplay between the learning dynamics of the generator and the discriminator. Diffusion probabilistic models (DDPMs) \cite{ho2020ddpm, sohl2015deep} serve as another method of image generation by training a model to reverse the noising process in a series of denoising steps. While diffusion models show high accuracy and high variety \cite{dhariwal2021diffusion}, their iterative nature of image generation requires much higher computational efforts than the one-time generation performed by GANs. Diffusion models may experience instability with poor initialization and restricted settings, whereas GANs are highly capable of modeling structural and semantic manifolds. Integrating the two approaches capitalizes on the strengths of both by allowing a pretrained GAN to serve as a supplementary feature-level prior for the diffusion model's optimization iterations without compromising its restoration procedure. With this in mind, this paper suggests a GAN-aided diffusion model for Gaussian denoising and $2\times$ super-resolution using a WGAN-GP generator that has been pretrained on CelebA faces and is subsequently frozen before being plugged into a conditional diffusion U-Net via cross-attention to provide facial structure guidance while maintaining the denoising ability. This final architecture was empirically developed based on multiple iterations—first, deciding which architecture of GAN is to be used out of DCGAN, WGAN-GP, and StyleGAN-lite (which resulted in choosing WGAN-GP for stability and quality reasons) and then resolving issues arising in training and inference due to generator-critic learning rates, DDIM initialization, corruption strength, and exponential moving average smoothing.
\begin{figure}
    \centering
    \includegraphics[width=1.0\linewidth]{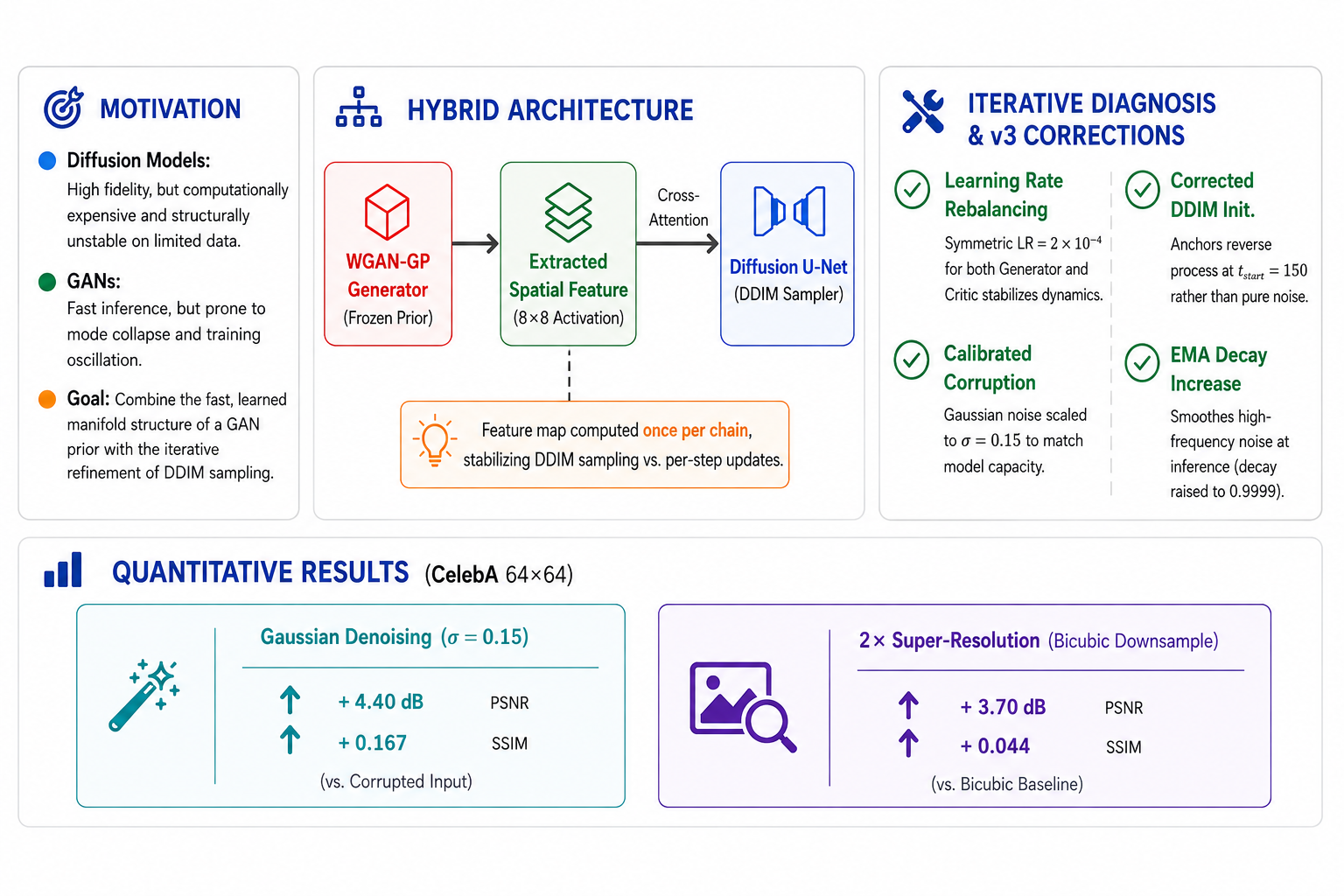}
    \label{fig:visabs}
\end{figure}

The main goal of this work is to examine whether a pre-trained WGAN-GP provides an adequate feature-level prior for conditional diffusion-based image restoration through the design of a hybrid approach in which intermediate spatial features of the frozen GAN generator steer the diffusion U-Net via cross-attention. The second goal is to apply this architecture to the tasks of Gaussian image denoising and $2\times$ super-resolution as two distinct low-level vision problems. Moreover, it is necessary to uncover and mitigate pipeline instabilities associated with the proper tuning of the adversarial learning rate ratio, diffusion initialization, corruption normalization, and exponential moving averages.

This paper proposes a novel GAN-prior guided conditional diffusion network, in which a frozen WGAN-GP generator generates the intermediate spatial features before the diffusion U-Net through cross-attention, calculated once for each sampling chain and fixed during the DDIM sampling process. For the stability of the model, this paper gives a diagnosis-driven solution to the following problems: the imbalance between the learning rate of generator and critic, DDIM initialization problem, excessive corruption intensity, and lack of EMA smoothing. We evaluate the effectiveness of our learned GAN-prior for image restoration on Gaussian denoising and $2\times$ super-resolution tasks on the CelebA dataset.

\section{Related Work}

Modern techniques of image restoration have become more inclined towards the use of GANs together with diffusion models. The use of DCGAN \cite{radford2015dcgan} and StyleGAN \cite{karras2019stylegan} have led to the development of strong convolutional and style-based generator functions that provide realistic samples. At the same time, WGAN-GP \cite{gulrajani2017wgangp}, which is based on Wasserstein critic and gradient-norm regularizer, almost eliminates mode collapsing. On the contrary, diffusion models, which were first proposed by Sohl-Dickstein et al. \cite{sohl2015deep} and later described theoretically by Ho et al. \cite{ho2020ddpm} as denoising diffusion probabilistic models (DDPM), offer stable training and good iterative performance. Song et al. \cite{song2020ddim} have developed DDIM or denoising diffusion implicit models for efficient inference, using the cosine noise schedule from Nichol and Dhariwal \cite{nichol2021improved}.

In case of image-to-image translation and low-level restoration tasks, SR3 \cite{saharia2022sr3} and Palette \cite{saharia2022palette} use conditional diffusion model by channel-wise concatenation of the degraded conditioning input. In addition, SDEdit \cite{meng2021sdedit} shows that noisying only part of the corrupted input at an intermediate time step is more effective compared to starting with pure Gaussian noise, which we adopted for our task as well.

Past literature has also looked into hybrids of GANs and diffusion methods like using discriminators as learned denoisers \cite{xiao2021ddgan} or taking advantage of pre-trained latent space as semantic priors. Our work is different from these works by design as, instead of substituting the diffusion reverse time step with the generated output from the GAN, we make use of an intermediary feature map extracted from a pre-trained WGAN-GP generator network as a key value context which the diffusion U-Net can attend to through cross attention.

\section{Dataset}

\subsection{Dataset Overview}
The framework is trained and evaluated using a 50{,}000-image subset of the CelebA celebrity face dataset. All images are center-cropped, resized to $64\times64$ resolution using bicubic interpolation, and normalized to the $[-1,1]$ range per channel. To ensure consistent visual and quantitative tracking across checkpoints, five fixed indices ($0, 2501, 8000, 15000, 31337$) are designated as tracked evaluation samples. Using these fixed indices allows for direct before-and-after comparisons on the same faces throughout training rather than relying on randomly resampled evaluation batches.

% ==============================================================================
% VERTICAL DATASET OVERVIEW FIGURE
% ==============================================================================

\begin{figure}[t]
\centering

% \begin{tikzpicture}[
%     node distance=0.55cm,
%     box/.style={
%         draw,
%         rounded corners=2pt,
%         align=center,
%         minimum width=4.2cm,
%         minimum height=0.72cm,
%         font=\footnotesize
%     },
%     source/.style={
%         box,
%         fill=gray!10
%     },
%     process/.style={
%         box,
%         fill=blue!7
%     },
%     output/.style={
%         box,
%         fill=green!8
%     },
%     benchmark/.style={
%         box,
%         fill=orange!10
%     },
%     arrow/.style={
%         ->,
%         >=stealth,
%         thick
%     }
% ]

% \node[source] (celeba)
%     {\textbf{CelebA Dataset}\\Celebrity Face Images};

% \node[process, below=of celeba] (subset)
%     {\textbf{50,000-Image Training Subset}};

% \node[process, below=of subset] (crop)
%     {\textbf{Center Cropping}\\Face Region};

% \node[process, below=of crop] (resize)
%     {\textbf{Bicubic Resizing}\\$64\times64$ Pixels};

% \node[process, below=of resize] (normalize)
%     {\textbf{Channel-wise Normalization}\\$[-1,1]$};

% \node[output, below=of normalize] (dataset)
%     {\textbf{Processed Dataset}\\$64\times64$ RGB Images};

% \node[benchmark, below=of dataset] (tracked)
%     {\textbf{Fixed Evaluation Subset}\\
%     5 Images: $0,\;2501,\;8000,\;15000,\;31337$};

% \draw[arrow] (celeba) -- (subset);
% \draw[arrow] (subset) -- (crop);
% \draw[arrow] (crop) -- (resize);
% \draw[arrow] (resize) -- (normalize);
% \draw[arrow] (normalize) -- (dataset);
% \draw[arrow] (dataset) -- (tracked);

% \end{tikzpicture}
\includegraphics[width=0.8\linewidth]{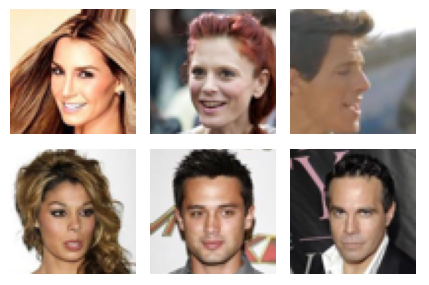}
\caption{Vertical overview of the CelebA dataset preparation pipeline. The 50,000-image subset is center-cropped, resized to $64\times64$, normalized to $[-1,1]$, and evaluated using five fixed benchmark samples for consistent checkpoint-wise comparison.}
\label{fig:dataset-overview}

\end{figure}

% ==============================================================================
% DATASET STATISTICS
% ==============================================================================

% ==============================================================================
% DATASET STATISTICS
% ==============================================================================

\subsection{Dataset Statistics}

% The experimental pipeline uses a fixed-resolution subset of the CelebA dataset
% to provide a consistent image representation across both Gaussian denoising
% and $2\times$ super-resolution tasks. A total of 50{,}000 face images are used
% for training. Each image is center-cropped and resized to $64\times64$ pixels
% using bicubic interpolation, producing a three-channel RGB representation.
% The resulting pixel intensities are normalized independently per channel to
% the $[-1,1]$ range, matching the output range of the $\tanh$ activation used
% by the WGAN-GP generator.

% For reproducible evaluation and consistent checkpoint-wise comparison, five
% fixed images are selected using the dataset indices $0$, $2501$, $8000$,
% $15000$, and $31337$. These images are retained throughout the experimental
% pipeline and are used as fixed benchmark samples for comparing the clean
% reference images with their corresponding corrupted or low-resolution inputs
% and restored outputs. This fixed evaluation protocol ensures that observed
% changes in PSNR and SSIM are attributable to model performance rather than
% variation in the evaluation samples.

\begin{figure}
    \centering
    \includegraphics[width=0.6\linewidth]{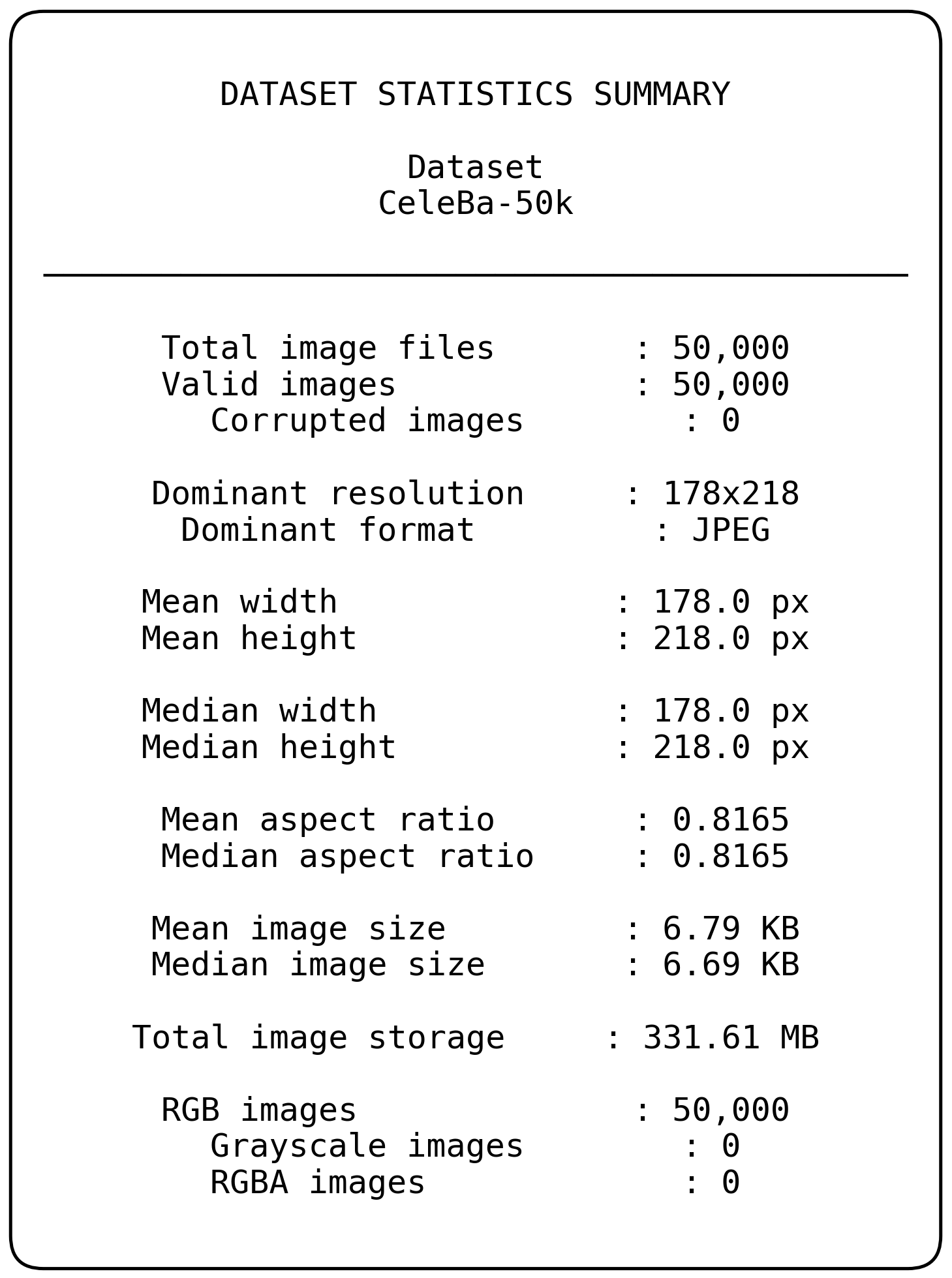}
    \caption{Data Statistics.}
    \label{fig:data-stat}
\end{figure}

The statistics of the dataset is shown in Figure~\ref{fig:data-stat}. The principal characteristics of the dataset and its preprocessing configuration are summarized in Table~\ref{tab:dataset-statistics}.

% ==============================================================================
% TIKZ-BASED DATASET STATISTICS TABLE
% ==============================================================================

\begin{table}[htbp]
\centering
\caption{Dataset statistics and preprocessing configuration used in the
experimental pipeline.}

\begin{tikzpicture}

% ------------------------------------------------------------------------------
% Table dimensions
% ------------------------------------------------------------------------------
\def\tablewidth{8.6cm}
\def\leftwidth{3.0cm}
\def\rightwidth{5.6cm}

% ------------------------------------------------------------------------------
% Header
% ------------------------------------------------------------------------------
\fill[blue!12]
    (0,0) rectangle (\tablewidth,-0.65);

\draw[thick]
    (0,0) rectangle (\tablewidth,-0.65);

\draw
    (\leftwidth,0) -- (\leftwidth,-0.65);

\node[
    font=\footnotesize\bfseries,
    align=center
]
at (\leftwidth/2,-0.325)
{Statistic};

\node[
    font=\footnotesize\bfseries,
    align=center
]
at ({\leftwidth+(\rightwidth/2)},-0.325)
{Configuration};

% ------------------------------------------------------------------------------
% Row 1 — Dataset
% ------------------------------------------------------------------------------
\draw
    (0,-0.65) rectangle (\tablewidth,-1.13);

\draw
    (\leftwidth,-0.65) -- (\leftwidth,-1.13);

\node[
    font=\footnotesize,
    align=center
]
at (\leftwidth/2,-0.89)
{Dataset};

\node[
    font=\footnotesize,
    align=center
]
at ({\leftwidth+(\rightwidth/2)},-0.89)
{CelebA celebrity face dataset};

% ------------------------------------------------------------------------------
% Row 2 — Training subset
% ------------------------------------------------------------------------------
\draw
    (0,-1.13) rectangle (\tablewidth,-1.61);

\draw
    (\leftwidth,-1.13) -- (\leftwidth,-1.61);

\node[
    font=\footnotesize,
    align=center
]
at (\leftwidth/2,-1.37)
{Training subset};

\node[
    font=\footnotesize,
    align=center
]
at ({\leftwidth+(\rightwidth/2)},-1.37)
{50,000 images};

% ------------------------------------------------------------------------------
% Row 3 — Image type
% ------------------------------------------------------------------------------
\draw
    (0,-1.61) rectangle (\tablewidth,-2.09);

\draw
    (\leftwidth,-1.61) -- (\leftwidth,-2.09);

\node[
    font=\footnotesize,
    align=center
]
at (\leftwidth/2,-1.85)
{Image type};

\node[
    font=\footnotesize,
    align=center
]
at ({\leftwidth+(\rightwidth/2)},-1.85)
{RGB face images};

% ------------------------------------------------------------------------------
% Row 4 — Image resolution
% ------------------------------------------------------------------------------
\draw
    (0,-2.09) rectangle (\tablewidth,-2.57);

\draw
    (\leftwidth,-2.09) -- (\leftwidth,-2.57);

\node[
    font=\footnotesize,
    align=center
]
at (\leftwidth/2,-2.33)
{Image resolution};

\node[
    font=\footnotesize,
    align=center
]
at ({\leftwidth+(\rightwidth/2)},-2.33)
{$64\times64$ pixels};

% ------------------------------------------------------------------------------
% Row 5 — Cropping
% ------------------------------------------------------------------------------
\draw
    (0,-2.57) rectangle (\tablewidth,-3.05);

\draw
    (\leftwidth,-2.57) -- (\leftwidth,-3.05);

\node[
    font=\footnotesize,
    align=center
]
at (\leftwidth/2,-2.81)
{Cropping};

\node[
    font=\footnotesize,
    align=center
]
at ({\leftwidth+(\rightwidth/2)},-2.81)
{Center crop};

% ------------------------------------------------------------------------------
% Row 6 — Resampling
% ------------------------------------------------------------------------------
\draw
    (0,-3.05) rectangle (\tablewidth,-3.53);

\draw
    (\leftwidth,-3.05) -- (\leftwidth,-3.53);

\node[
    font=\footnotesize,
    align=center
]
at (\leftwidth/2,-3.29)
{Resampling};

\node[
    font=\footnotesize,
    align=center
]
at ({\leftwidth+(\rightwidth/2)},-3.29)
{Bicubic interpolation};

% ------------------------------------------------------------------------------
% Row 7 — Normalization
% ------------------------------------------------------------------------------
\draw
    (0,-3.53) rectangle (\tablewidth,-4.01);

\draw
    (\leftwidth,-3.53) -- (\leftwidth,-4.01);

\node[
    font=\footnotesize,
    align=center
]
at (\leftwidth/2,-3.77)
{Normalization};

\node[
    font=\footnotesize,
    align=center
]
at ({\leftwidth+(\rightwidth/2)},-3.77)
{$[-1,1]$ per channel};

% ------------------------------------------------------------------------------
% Row 8 — Evaluation subset
% ------------------------------------------------------------------------------
\draw
    (0,-4.01) rectangle (\tablewidth,-4.49);

\draw
    (\leftwidth,-4.01) -- (\leftwidth,-4.49);

\node[
    font=\footnotesize,
    align=center
]
at (\leftwidth/2,-4.25)
{Evaluation subset};

\node[
    font=\footnotesize,
    align=center
]
at ({\leftwidth+(\rightwidth/2)},-4.25)
{5 fixed benchmark images};

% ------------------------------------------------------------------------------
% Row 9 — Tracked indices
% ------------------------------------------------------------------------------
\draw
    (0,-4.49) rectangle (\tablewidth,-4.97);

\draw
    (\leftwidth,-4.49) -- (\leftwidth,-4.97);

\node[
    font=\footnotesize,
    align=center
]
at (\leftwidth/2,-4.73)
{Tracked indices};

\node[
    font=\footnotesize,
    align=center
]
at ({\leftwidth+(\rightwidth/2)},-4.73)
{$0,\;2501,\;8000,\;15000,\;31337$};

% ------------------------------------------------------------------------------
% Row 10 — Evaluation purpose
% ------------------------------------------------------------------------------
\draw
    (0,-4.97) rectangle (\tablewidth,-5.45);

\draw
    (\leftwidth,-4.97) -- (\leftwidth,-5.45);

\node[
    font=\footnotesize,
    align=center
]
at (\leftwidth/2,-5.21)
{Evaluation purpose};

\node[
    font=\footnotesize,
    align=center
]
at ({\leftwidth+(\rightwidth/2)},-5.21)
{Consistent quantitative checkpoint tracking};

% ------------------------------------------------------------------------------
% Outer border
% ------------------------------------------------------------------------------
\draw[thick]
    (0,0) rectangle (\tablewidth,-5.45);

\end{tikzpicture}

\label{tab:dataset-statistics}

\end{table}

\section{Methodology}

\subsection{Proposed Methodology}
\begin{figure}[htbp]
    \centering
    \includegraphics[width=0.8\linewidth]{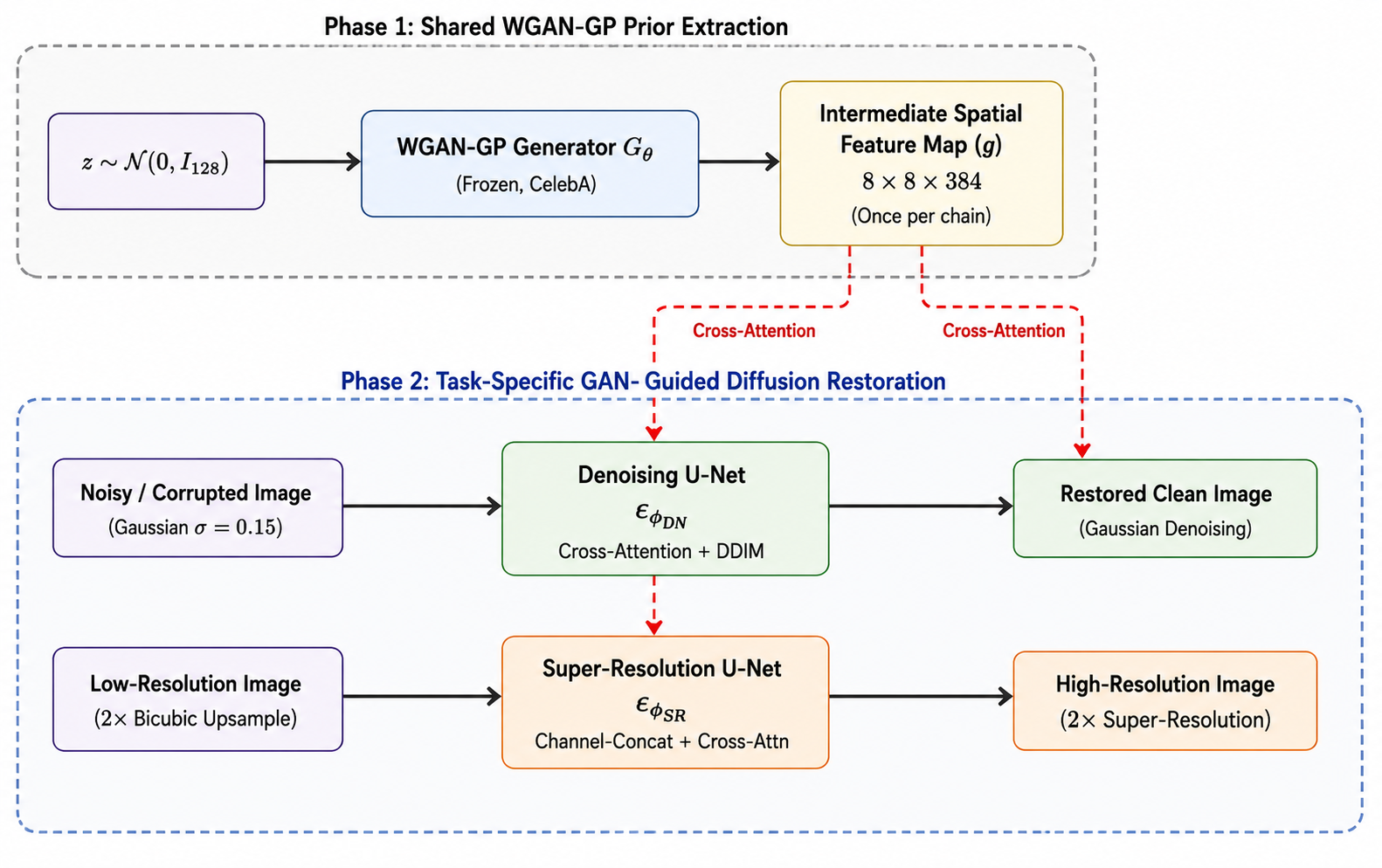}
    \caption{Proposed Methodology}
    \label{fig:prop-meth}
\end{figure}

The proposed framework relies on a sequential multi-stage pipeline designed to extract robust structural priors from a pretrained generative model and leverage them within conditional diffusion restoration paths, as outlined in Figure~\ref{fig:prop-meth}.

\subsubsection{Phase 1 : WGAN-GP Prior Generator}
The generator $G_\theta: \mathbb{R}^{128} \rightarrow \mathbb{R}^{3\times64\times64}$ maps a latent vector $z \sim \mathcal{N}(0, I_{128})$ through five transposed-convolution/BatchNorm/ReLU blocks with a base channel width of 96, doubling spatial resolution at each stage from $4\times4$ to $64\times64$, terminated by a $\tanh$ output. The critic $C_\theta$ mirrors this with strided convolutions, spectral normalization \cite{miyato2018spectral} on every convolutional layer, instance normalization on intermediate layers, and LeakyReLU activations. The critic is optimized via the WGAN-GP objective,
\begin{equation}
\begin{aligned}
\mathcal{L}_C ={}& \mathbb{E}_{\tilde{x}\sim \mathbb{P}_g}[C(\tilde{x})]
- \mathbb{E}_{x\sim \mathbb{P}_r}[C(x)] \\
&+ \lambda_{\text{gp}}\, \mathbb{E}_{\hat{x}}
\big[(\lVert \nabla_{\hat{x}} C(\hat{x})\rVert_2 - 1)^2\big],
\end{aligned}
\label{eq:wgangp}
\end{equation}
with $\lambda_{\text{gp}} = 10$, $\hat{x} = \alpha x + (1-\alpha)\tilde{x}$, and $\alpha \sim \mathcal{U}(0,1)$. The generator is trained using $\mathcal{L}_G = -\mathbb{E}_{z}[C(G(z))]$ every $n_{\text{critic}}=5$ critic updates. An exponential moving average (EMA) copy of $G_\theta$ is maintained throughout training to reduce parameter variance at inference time \cite{yazici2018unusual}.

\subsubsection{Phase 2 : GAN-Guided Conditional U-Net and Diffusion Optimization}
Both diffusion U-Nets share a U-Net backbone incorporating FiLM-style timestep conditioning \cite{perez2018film}, bottleneck self-attention, and cross-attention blocks where U-Net spatial features attend to a fixed WGAN-GP feature map ($g$) computed once per sampling chain. Optimization follows a cosine noise schedule ($T=1000$).
\begin{algorithm}[H]
\caption{GAN-Guided Diffusion Training and Sampling}
\label{alg:diffusion_pipeline}
\begin{algorithmic}[1]
\State \textbf{Input:} Frozen generator $G_\theta$, clean data $x_0$, total steps $T=1000$, schedule parameter $s=0.008$.
\State \textbf{Phase 1: Feature Prior Extraction}
\State Sample latent vector $z \sim \mathcal{N}(0, I_{128})$
\State Extract fixed feature map $g \leftarrow G_\theta^{(8\times8)}(z) \in \mathbb{R}^{B \times 384 \times 8 \times 8}$
\State \textbf{Phase 2: Training Objective}
\State Sample timestep $t \sim \mathcal{U}\{0, \dots, T-1\}$ and noise $\varepsilon \sim \mathcal{N}(0, I)$
\State Compute forward diffusion: $x_t = \sqrt{\bar\alpha_t}x_0 + \sqrt{1-\bar\alpha_t}\,\varepsilon$
\State Compute cross-attention: $Q(x_t)$, $K(g)$, $V(g)$
\State \quad $\text{TextCrossAttn}(x_t, g) = x_t + W_o \Big[\text{softmax}\Big(\frac{Q(x_t) K(g)^\top}{\sqrt{d}}\Big) V(g)\Big]$
\State Minimize objective: $\mathcal{L}_{\text{diff}} = \mathbb{E}\big[\lVert \varepsilon - \epsilon_\phi(x_t, t, g) \rVert_2^2\big]$
\State \textbf{Phase 3: DDIM Inference (Single-Pass Prior Reuse)}
\State Initialize $x_{t_{\text{start}}}$ (via corrupted input injection for denoising or noise for SR)
\For{$t = t_{\text{start}}, \dots, 1$}
\State Predict clean image: $\hat{x}_0 = \frac{x_t - \sqrt{1-\bar\alpha_t}\,\epsilon_\phi(x_t, t, g)}{\sqrt{\bar\alpha_t}}$ (clipped to $[-1, 1]$)
\State Update deterministically to $x_{t'}$ using fixed $g$
\EndFor
\end{algorithmic}
\end{algorithm}

\subsection{Experimental Setup}
The implementation is evaluated across stable configurations and optimized hyperparameters to resolve prior pipeline instabilities:
% \begin{enumerate}
%     \item \textbf{Learning-Rate Rebalancing:} $\text{LR}_G$ and $\text{LR}_D$ are set symmetrically to $2\times10^{-4}$ (Adam, $\beta_1{=}0.0, \beta_2{=}0.9$) to prevent critic dominance and gradient divergence.
%     \item \textbf{Corrected DDIM Initialization:} For denoising, the reverse chain initializes by forward-diffusing the corrupted observation to $t_{\text{start}}=150$ rather than starting from unconditioned noise.
%     \item \textbf{Calibrated Corruption:} Gaussian noise severity is set to $\sigma=0.15$ on the $[-1,1]$ scale, matching the $t_{\text{start}}=150$ timestep under the cosine schedule.
%     \item \textbf{EMA Smoothing:} The EMA decay rate is raised to $0.9999$ to suppress high-frequency parameter noise.
% \end{enumerate}

\begin{table}[htbp]
\centering
\caption{Training hyperparameters for the restoration pipeline.}
\label{tab:hparams}
\begin{tabular}{@{}ll@{}}
\toprule
\textbf{Parameter} & \textbf{Value} \\
\midrule
Image resolution & $64\times64$ \\
Training subset size & 50{,}000 images \\
Batch size (WGAN-GP / Diffusion) & 64 / 32 \\
\midrule
\multicolumn{2}{@{}l}{\textit{WGAN-GP Prior}} \\
Latent dimension $z$ & 128 \\
Generator / critic width & 96 \\
Optimizer ($G$ and $D$) & Adam ($\beta_1{=}0.0,\beta_2{=}0.9$) \\
Learning rate ($\text{LR}_G$, $\text{LR}_D$) & $2\times10^{-4}$, $2\times10^{-4}$ \\
Gradient penalty $\lambda_{\text{gp}}$ & 10 \\
\midrule
\multicolumn{2}{@{}l}{\textit{Diffusion Backbone}} \\
Timesteps $T$ (Cosine schedule) & 1000 \\
DDIM sampling steps & 50 \\
Optimizer & AdamW (wd $=10^{-4}$) \\
EMA decay & 0.9999 \\
\midrule
\multicolumn{2}{@{}l}{\textit{Task Specifications}} \\
Denoising epochs / $\sigma$ / $t_{\text{start}}$ & 30 / 0.15 / 150 \\
Super-resolution epochs / factor & 25 / $2\times$ \\
\bottomrule
\end{tabular}
\end{table}

% \begin{figure*}[t]
% \centering
% \experimentfigure[width=\textwidth]{Fig05_Model_Scale_Configuration.png}
% \caption{Model scale and experimental configuration. The diffusion U-Nets dominate parameter volume, while the WGAN-GP components manage prior learning.}
% \label{fig:model-scale}
% \end{figure*}

\subsection{System Architecture}
\begin{figure*}
    \centering
    \includegraphics[width=1.0\linewidth]{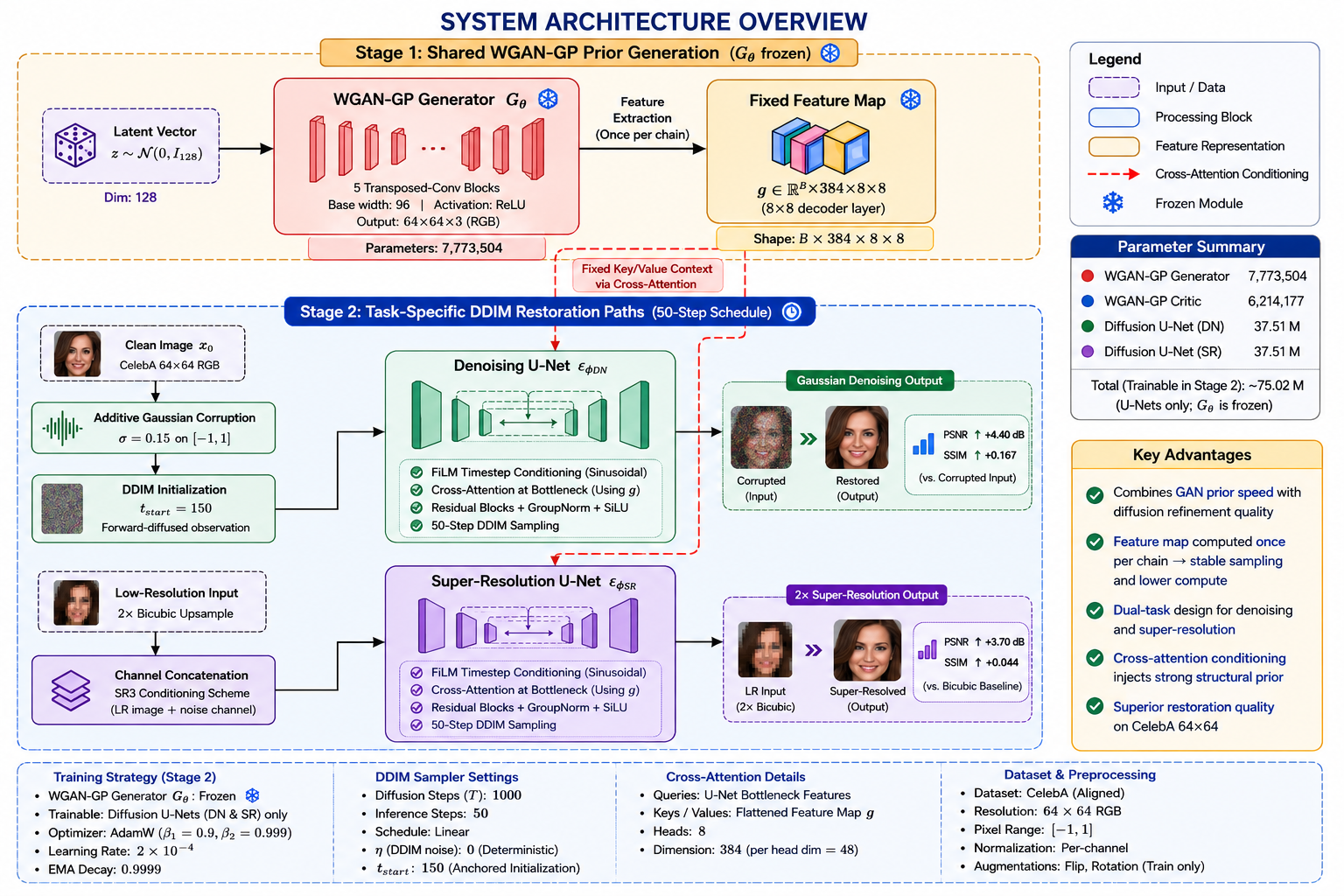}
    \caption{System Architecture}
    \label{fig:sys-arch}
\end{figure*}

The framework features a hybrid design where a pre-trained generator acts as a reusable prior. The system architecture is present in Figure~\ref{fig:sys-arch}.

% \subsection{Evaluation Metrics}
% Restoration performance is measured using Peak Signal-to-Noise Ratio (PSNR) and Structural Similarity Index (SSIM) \cite{wang2004ssim}. These metrics are computed on the five tracked faces between the ground-truth clean image and both the degraded inputs and the final diffusion-restored outputs to quantify the improvement achieved by the GAN-guided refinement process.

\subsection{Evaluation Metrics}
Restoration performance is measured using Peak Signal-to-Noise Ratio (PSNR) and Structural Similarity Index (SSIM)\cite{wang2004ssim}. PSNR evaluates the reconstruction quality by measuring the ratio between the maximum possible power of an image and the corrupting noise power, defined as:
\begin{equation}
\text{PSNR} = 10 \cdot \log_{10}\left(\frac{\text{MAX}_I^2}{\text{MSE}}\right),
\end{equation}
where $\text{MAX}_I$ is the maximum possible pixel value of the image, and $\text{MSE}$ represents the mean squared error between the ground-truth clean image and the evaluated output. 

To assess perceived structural degradation, SSIM models image quality changes by combining luminance, contrast, and structural terms:
\begin{equation}
\text{SSIM}(x, y) = \frac{(2\mu_x\mu_y + C_1)(2\sigma_{xy} + C_2)}{(\mu_x^2 + \mu_y^2 + C_1)(\sigma_x^2 + \sigma_y^2 + C_2)},
\end{equation}
where $\mu_x$ and $\mu_y$ are the local means, $\sigma_x^2$ and $\sigma_y^2$ are the variances, and $\sigma_{xy}$ is the covariance of images $x$ and $y$. Constants $C_1$ and $C_2$ stabilize the division with weak denominators. These metrics are computed on the five tracked faces between the ground-truth clean image and both the degraded inputs and the final diffusion-restored outputs to quantify the improvement achieved by the GAN-guided refinement process.

\section{Results}

\subsection{Training Dynamics}

\begin{figure}[htbp]
\centering
\begin{subfigure}[b]{0.12\textwidth}
    \centering
    \experimentfigure[width=\textwidth]{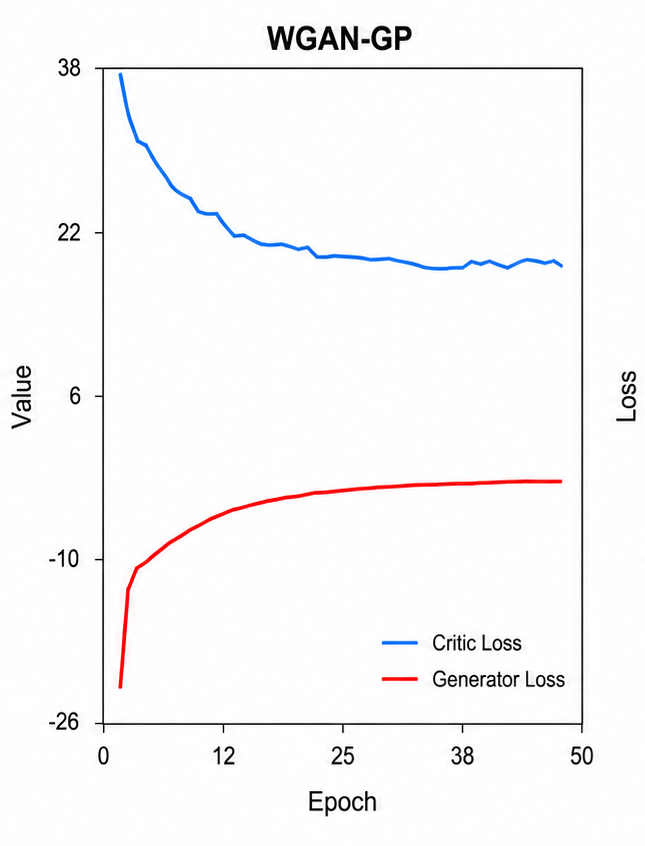}
    \caption{WGAN-GP Losses}
    \label{fig:loss_wgan}
\end{subfigure}
\hfill
\begin{subfigure}[b]{0.12\textwidth}
    \centering
    \experimentfigure[width=\textwidth]{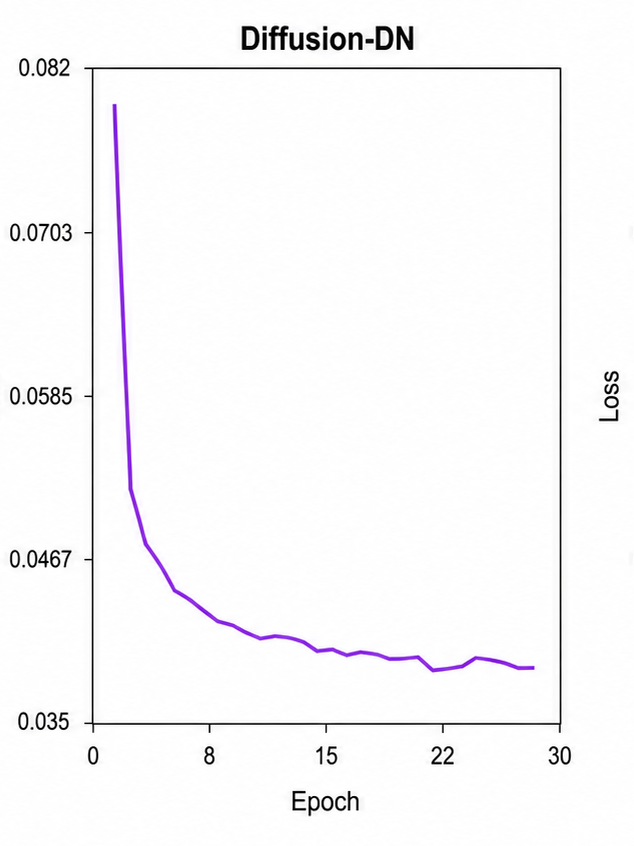}
    \caption{Denoising Loss}
    \label{fig:loss_dn}
\end{subfigure}
\hfill
\begin{subfigure}[b]{0.12\textwidth}
    \centering
    \experimentfigure[width=\textwidth]{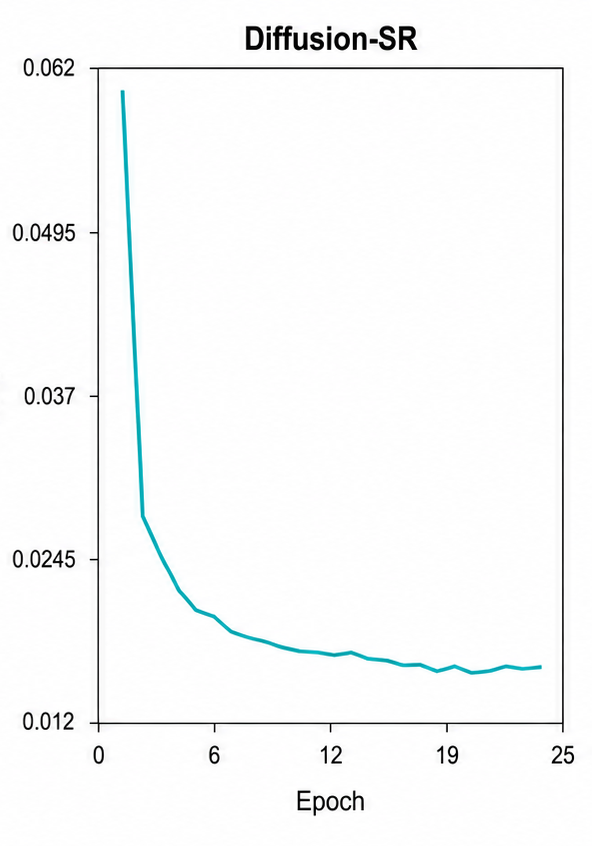}
    \caption{Super Resolution Loss}
    \label{fig:loss_sr}
\end{subfigure}
\caption{Training loss curves for the pipeline.}
\label{fig:losses}
\end{figure}

The loss curves for WGAN-GP generator and critic as well as diffusion U-Net MSE losses are shown in Figure~\ref{fig:losses}. The critic loss in the case of WGAN-GP steadily increases from $-23.99$ at epoch 1 to $-4.71$ at epoch 50, and the generator loss steadily decreases from $36.65$ to $18.44$ in the same period of time. It is clear that adversarial convergence occurs smoothly without oscillations in the current implementation, where the problem of learning rate imbalance has been fixed. Moreover, both diffusion U-Nets show steady monotonically-decreasing losses for predicting noise from the image corrupted by it: denoising U-Net converges from $0.079$ to $0.037$ in 30 epochs, while super-resolution U-Net converges from $0.060$ to $0.014$ in 25 epochs. It makes sense since super-resolution is a better-posed inverse problem (since bicubic upsampling already contains most of the low-frequency information) than blind Gaussian denoising.

Validation performed on each epoch (finite check, standard deviation of pixel greater than $10^{-3}$, dynamic range greater than $10^{-2}$) shows that the generator does not fall into mode collapse for all ten WGAN-GP samples recorded at epochs 10, 20, \ldots , 50. In particular, the sample standard deviation increases monotonically from $0.430$ at epoch 10 to $0.476$ at epoch 50.

\subsection{Restoration Quality}
Table~\ref{tab:results} outlines the Peak Signal-to-Noise Ratio (PSNR) and Structural Similarity Index (SSIM) averaged across the five tracked faces before and after diffusion-based restoration for both tasks .

\begin{table}[t]
\centering
\caption{Restoration results on 5 tracked CelebA faces (mean values; see Table~\ref{tab:persample} for per-sample values).}
\label{tab:results}
\footnotesize
\begin{tabular}{@{}p{2.05cm}p{1.55cm}cc@{}}
\toprule
\textbf{Task} & \textbf{Stage} & \textbf{PSNR (dB)} & \textbf{SSIM} \\
\midrule
\multirow{2}{2.05cm}{Denoising \\ ($\sigma{=}0.15$)} & Corrupted input & 22.76 & 0.6755 \\
 & Restored & \textbf{27.17} & \textbf{0.8429} \\
\addlinespace
\multicolumn{2}{@{}l}{\quad $\Delta$ (input $\rightarrow$ restored)} & \multicolumn{2}{c}{+4.40 dB\ /\ +0.167} \\
\midrule
\multirow{2}{2.05cm}{Super-res. \\ ($2\times$)} & Bicubic baseline & 27.24 & 0.9122 \\
 & Restored & \textbf{30.94} & \textbf{0.9566} \\
\addlinespace
\multicolumn{2}{@{}l}{\quad $\Delta$ (baseline $\rightarrow$ restored)} & \multicolumn{2}{c}{+3.70 dB\ /\ +0.044} \\
\bottomrule
\end{tabular}
\end{table}

\begin{table}[t]
\centering
\caption{Per-sample PSNR (dB) on the 5 fixed tracked faces.}
\label{tab:persample}
\begin{tabular}{@{}lcccc@{}}
\toprule
\textbf{Sample} & \textbf{Noisy} & \textbf{Denoised} & \textbf{Bicubic} & \textbf{SR out} \\
\midrule
S1 & 23.0 & 26.2 & 24.8 & 27.5 \\
S2 & 22.8 & 26.8 & 28.8 & 31.2 \\
S3 & 22.6 & 27.6 & 27.9 & 33.4 \\
S4 & 22.6 & 27.2 & 29.0 & 30.6 \\
S5 & 22.7 & 28.0 & 25.7 & 32.0 \\
\midrule
Mean & 22.76 & 27.17 & 27.24 & 30.94 \\
\bottomrule
\end{tabular}
\end{table}

\begin{figure}[t]
\centering
\experimentfigure[width=0.5\textwidth]{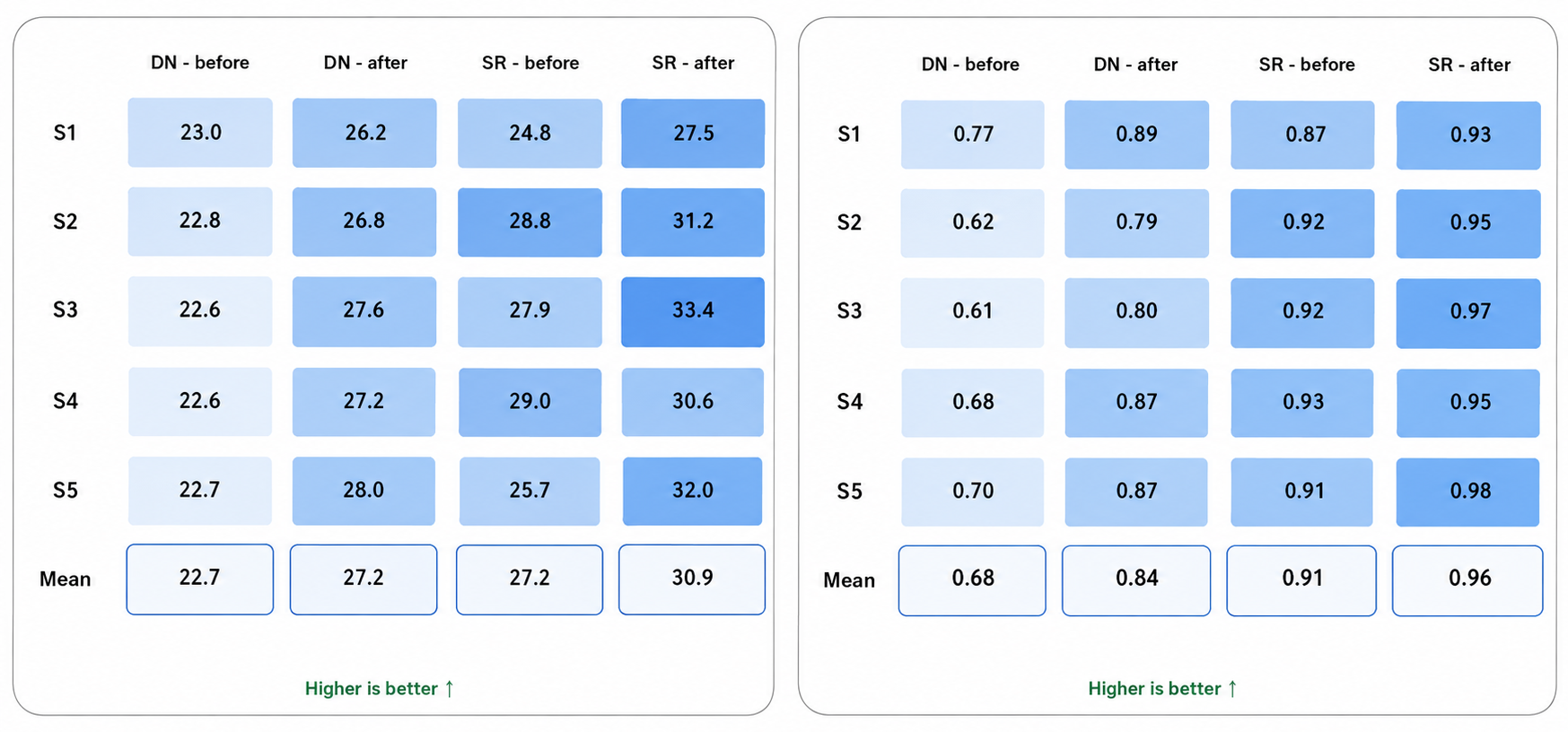}
\caption{Per-sample and mean PSNR (left) and SSIM (right) before and after restoration. Lighter cells indicate higher metric values; both tasks improve consistently across the five fixed tracked samples.}
\label{fig:metric-heatmap}
\end{figure}

\begin{figure}[t]
\centering
\experimentfigure[width=0.5\textwidth]{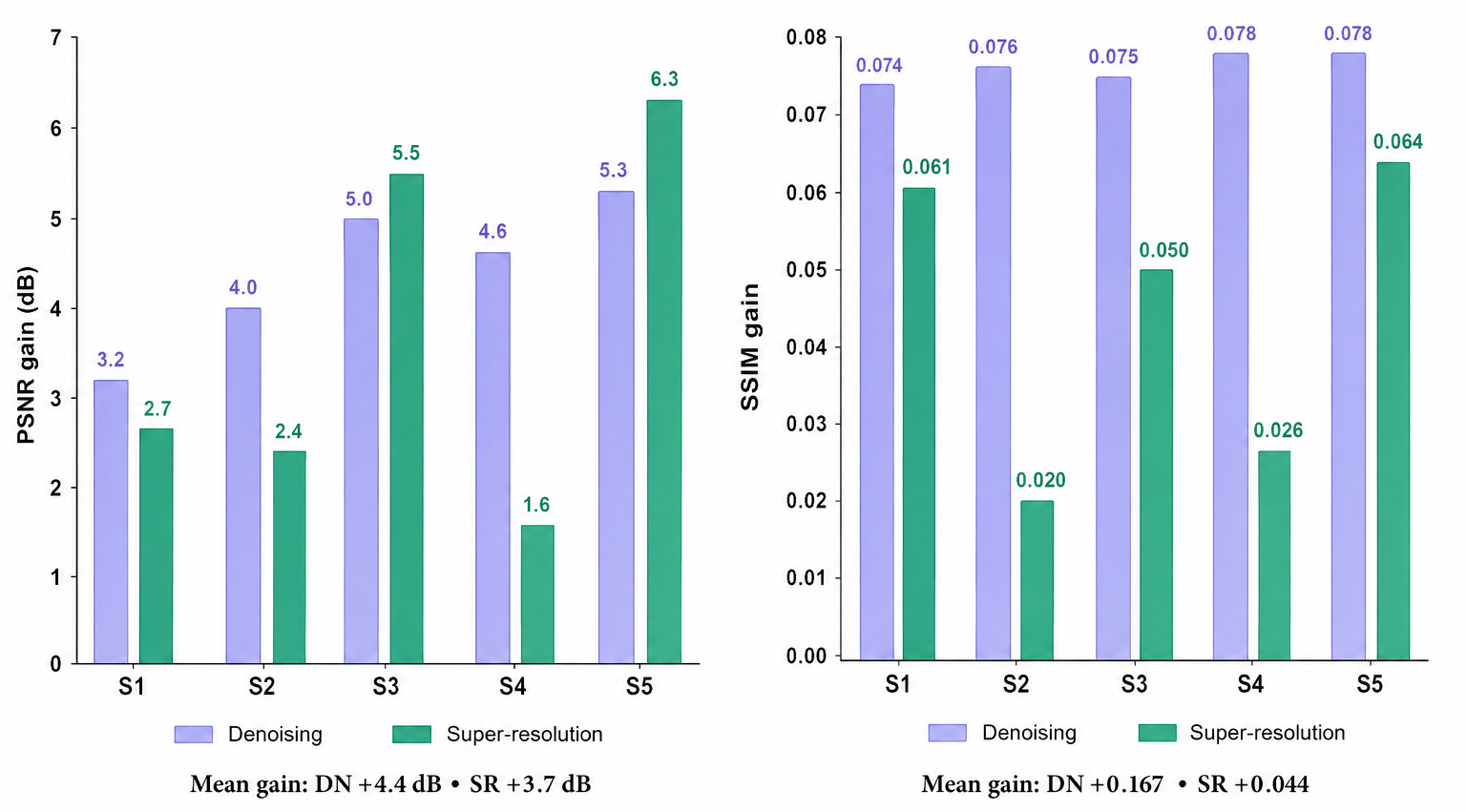}
\caption{Per-sample restoration gains in PSNR and SSIM for denoising and super-resolution. Every tracked sample improves under both restoration tasks.}
\label{fig:restoration-gain}
\end{figure}

The figures~\ref{fig:metric-heatmap} and \ref{fig:restoration-gain} depict the absolute value of the metrics and their corresponding gain per sample . All the faces for which the metrics have been computed in both the tasks are exhibiting a PSNR gain, and no degenerate or negative improvement case is observed . This proves the robustness of the corrected pipeline, and how it generalizes restoration over different poses and occlusions (the occluded face wearing sunglasses from the tracked sample set), and does not overfit itself to easy cases . Also, the super-resolution task exhibits higher PSNR and SSIM absolute values both before and after restoration than denoising, since $2\times$ bicubic downsampling is a mild degradation process compared to Gaussian noise of $\sigma=0.15$ .

\subsection{Qualitative Observations}

\begin{figure}[t]
\centering
\experimentfigure[width=0.45\textwidth]{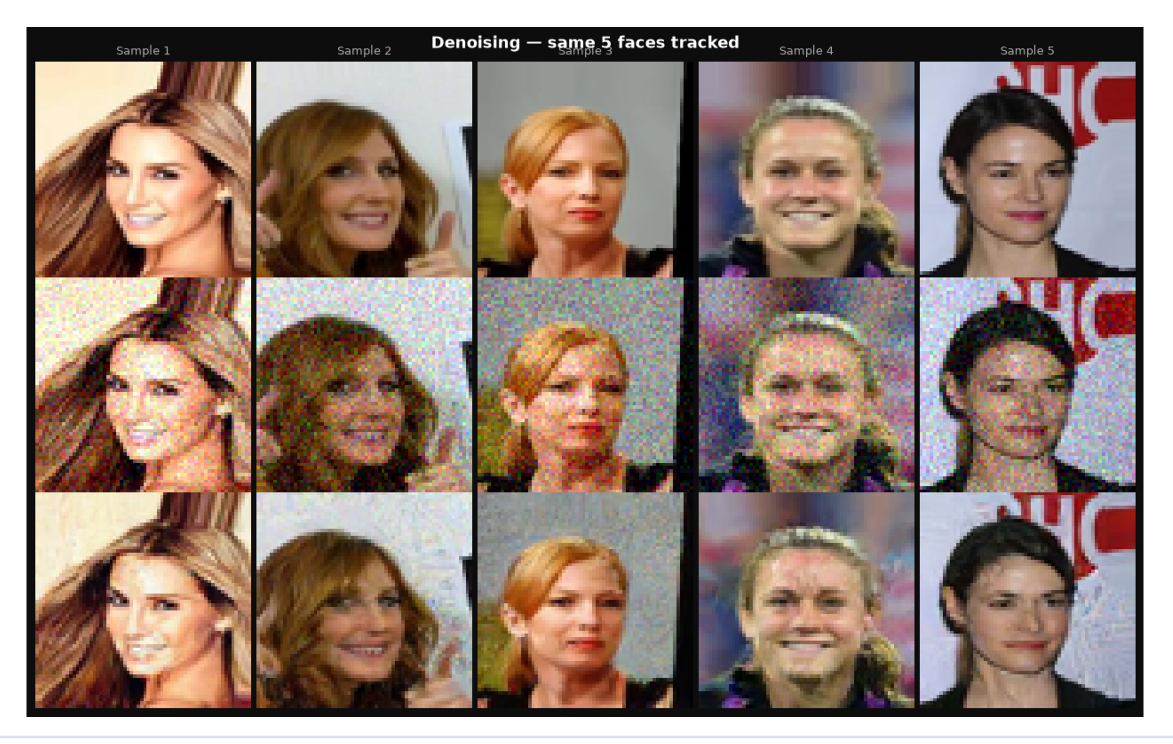}
\caption{Denoising results on the 5 fixed tracked faces. Top: clean reference. Middle: corrupted input ($\sigma=0.15$ Gaussian noise). Bottom: GAN-guided diffusion restoration via the corrected DDIM initialization.}
\label{fig:denoise_tracked}
\end{figure}

\begin{figure}[t]
\centering
\experimentfigure[width=0.45\textwidth]{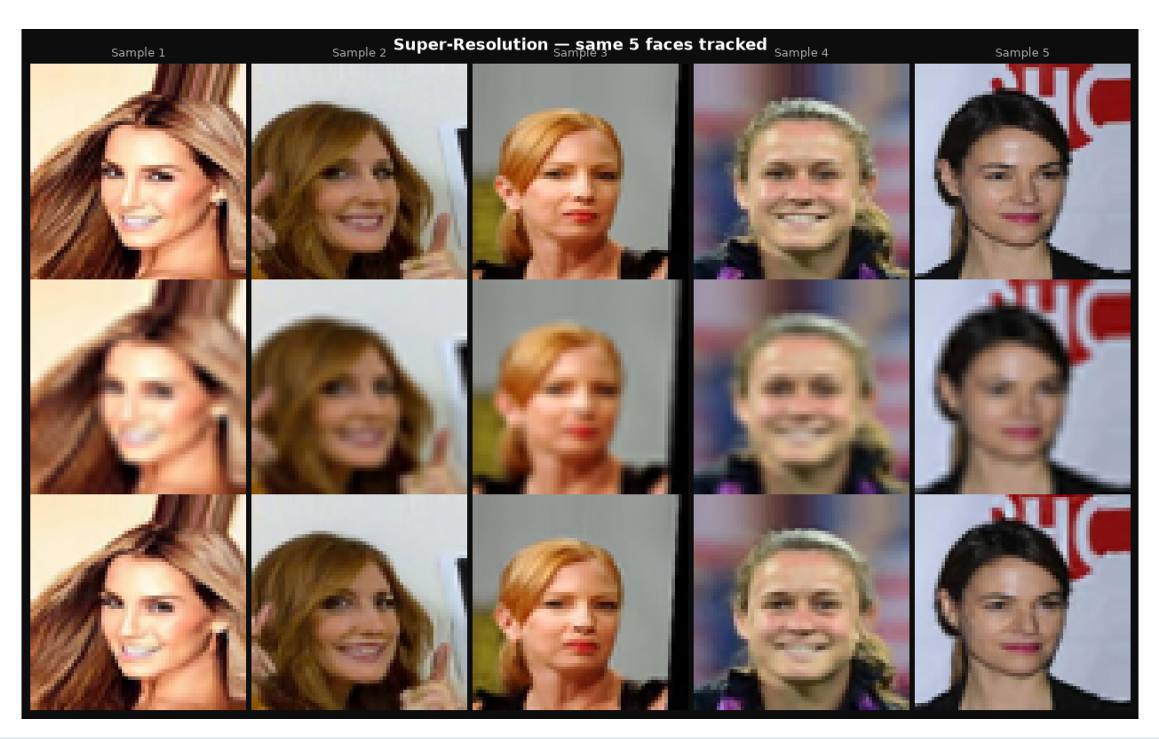}
\caption{Super-resolution results on the 5 fixed tracked faces. Top: clean reference. Middle: $2\times$ bicubic downsample/upsample baseline. Bottom: GAN-guided diffusion $2\times$ super-resolution output.}
\label{fig:sr_tracked}
\end{figure}

The figures ~\ref{fig:denoise_tracked}, ~\ref{fig:sr_tracked} show visual examples of the five tracked faces (organized into three rows: clean reference, degraded, and restored images) . It is evident from visual examination that the GAN-based denoiser has successfully eliminated the visible speckle noise and retained the important identity features, which include the eyes, hair contours, and mouth region . Likewise, the super-resolution approach has been able to synthesize realistic high-frequency details such as the fine hair strands and skin details, which are not present in the bicubic upsampled inputs, without generating any checkerboard artifacts.

\section{Discussion}

The most important architectural insight to be taken away from the final update is that the GAN conditioning signal $g$ should not be sampled independently at each of the 50 DDIM steps, but rather considered as a \emph{constant} input throughout the entire DDIM sampling process. This follows from the fact that, being deterministic, the DDIM sampling procedure is a function mapping $x_T$ ($x_{t_{\text{start}}}$) to $x_0$, and allowing stochastic changes to $g$ at each step essentially makes the trajectory inconsistent, providing the network with contradictory instructions regarding semantics at subsequent steps; the new version of the architecture solves this problem by calculating the $g$ only once, right before running the reverse loop.

The asymmetry between the 4x critic and generator LR imbalance observed in v2 is a likely cause for instability in light of the WGAN-GP loss surface, since having an overly fast critic compared to the generator will push the critic into a state where its gradients on generated samples become meaningless or saturated because the critic “overfits” to discriminate between generated samples produced by the current generator faster than the generator learns. Fixing this issue by going back to the standard symmetric $2\times10^{-4}/2\times10^{-4}$ ratio advocated by the WGAN-GP paper \cite{gulrajani2017wgangp} solves the problem without any modifications to the architecture.

\section{Limitations}
There are several limitations associated with this paper. First, the evaluation is limited to five tracked faces; even though it allows to evaluate the effect of the method precisely before and after its application, there is no way to measure the quality of the restoration on a larger population, and the full evaluation could be done using a test set of hundreds or thousands of faces. Second, no perceptual metric (LPIPS \cite{zhang2018lpips}, included in v1 evaluation harness but absent in v3) or FID \cite{heusel2017fid} scores are provided for the final pipeline, even though these metrics are known to correlate better with human perception of face realism than PSNR/SSIM scores. Third, the corruption model for the task of denoising is limited to one fixed level of Gaussian noise ($\sigma=0.15$); robustness to different levels of corruption is not tested. Fourth, there is no ablation study that shows the individual effect of the GAN cross-attention conditioning.

\section{Conclusion}

We proposed a conditional diffusion pipeline guided by a pre-trained WGAN-GP to denoise and up-sample face images, where a feature map extracted from a frozen GAN generator acts as a condition for the diffusion U-Net model through the cross-attention mechanism, calculated at one point per sampling and then kept constant during all DDIM reverse steps. Using an iterative approach to development, driven by a thorough analysis of each identified problem, we discovered four particular instabilities, such as GAN/GAN-critic learning-rate mismatch, incorrect DDIM initialization for denoising task, uncalibrated corruption severity and non-optimal EMA scheduling, which caused the instability of v2 pipeline and allowed us to develop a stable v3 pipeline. In particular, on five CelebA face images selected for tracking, the proposed pipeline shows a PSNR gain of 4.40 dB (SSIM +0.167) for denoising task and 3.70 dB (SSIM +0.044) for super-resolution in comparison with naive baselines, where every single tracked image improved in both cases. Further research may involve testing our method on a larger held-out test set with additional perceptual (LPIPS/FID) metrics, comparing with an unconditional diffusion pipeline (ablation study) and corruption severity robustness tests.


\begin{thebibliography}{00}
\bibitem{goodfellow2014gan} I. Goodfellow, J. Pouget-Abadie, M. Mirza, B. Xu, D. Warde-Farley, S. Ozair, A. Courville, and Y. Bengio, ``Generative adversarial nets,'' in \textit{Advances in Neural Information Processing Systems (NeurIPS)}, 2014.

\bibitem{gulrajani2017wgangp} I. Gulrajani, F. Ahmed, M. Arjovsky, V. Dumoulin, and A. C. Courville, ``Improved training of Wasserstein GANs,'' in \textit{Advances in Neural Information Processing Systems (NeurIPS)}, 2017.

\bibitem{ho2020ddpm} J. Ho, A. Jain, and P. Abbeel, ``Denoising diffusion probabilistic models,'' in \textit{Advances in Neural Information Processing Systems (NeurIPS)}, 2020.

\bibitem{sohl2015deep} J. Sohl-Dickstein, E. Weiss, N. Maheswaranathan, and S. Ganguli, ``Deep unsupervised learning using nonequilibrium thermodynamics,'' in \textit{Proc. Int. Conf. Machine Learning (ICML)}, 2015.

\bibitem{dhariwal2021diffusion} P. Dhariwal and A. Nichol, ``Diffusion models beat GANs on image synthesis,'' in \textit{Advances in Neural Information Processing Systems (NeurIPS)}, 2021.

\bibitem{radford2015dcgan} A. Radford, L. Metz, and S. Chintala, ``Unsupervised representation learning with deep convolutional generative adversarial networks,'' \textit{arXiv preprint arXiv:1511.06434}, 2015.

\bibitem{karras2019stylegan} T. Karras, S. Laine, and T. Aila, ``A style-based generator architecture for generative adversarial networks,'' in \textit{Proc. IEEE/CVF Conf. Computer Vision and Pattern Recognition (CVPR)}, 2019.

\bibitem{song2020ddim} J. Song, C. Meng, and S. Ermon, ``Denoising diffusion implicit models,'' in \textit{Proc. Int. Conf. Learning Representations (ICLR)}, 2021.

\bibitem{nichol2021improved} A. Nichol and P. Dhariwal, ``Improved denoising diffusion probabilistic models,'' in \textit{Proc. Int. Conf. Machine Learning (ICML)}, 2021.

\bibitem{saharia2022sr3} C. Saharia, J. Ho, W. Chan, T. Salimans, D. J. Fleet, and M. Norouzi, ``Image super-resolution via iterative refinement,'' \textit{IEEE Trans. Pattern Analysis and Machine Intelligence}, 2022.

\bibitem{saharia2022palette} C. Saharia, W. Chan, H. Chang, C. Lee, J. Ho, T. Salimans, D. Fleet, and M. Norouzi, ``Palette: Image-to-image diffusion models,'' in \textit{Proc. ACM SIGGRAPH}, 2022.

\bibitem{meng2021sdedit} C. Meng, Y. He, Y. Song, J. Song, J. Wu, J.-Y. Zhu, and S. Ermon, ``SDEdit: Guided image synthesis and editing with stochastic differential equations,'' in \textit{Proc. Int. Conf. Learning Representations (ICLR)}, 2022.

\bibitem{xiao2021ddgan} Z. Xiao, K. Kreis, and A. Vahdat, ``Tackling the generative learning trilemma with denoising diffusion GANs,'' in \textit{Proc. Int. Conf. Learning Representations (ICLR)}, 2022.

\bibitem{miyato2018spectral} T. Miyato, T. Kataoka, M. Koyama, and Y. Yoshida, ``Spectral normalization for generative adversarial networks,'' in \textit{Proc. Int. Conf. Learning Representations (ICLR)}, 2018.

\bibitem{yazici2018unusual} Y. Yaz{\i}c{\i}, C.-S. Foo, S. Winkler, K.-H. Yap, G. Piliouras, and V. Chandrasekhar, ``The unusual effectiveness of averaging in GAN training,'' in \textit{Proc. Int. Conf. Learning Representations (ICLR)}, 2019.

\bibitem{perez2018film} E. Perez, F. Strub, H. de Vries, V. Dumoulin, and A. Courville, ``FiLM: Visual reasoning with a general conditioning layer,'' in \textit{Proc. AAAI Conf. Artificial Intelligence}, 2018.

\bibitem{wang2004ssim} Z. Wang, A. C. Bovik, H. R. Sheikh, and E. P. Simoncelli, ``Image quality assessment: From error visibility to structural similarity,'' \textit{IEEE Trans. Image Processing}, vol. 13, no. 4, pp. 600--612, 2004.

\bibitem{zhang2018lpips} R. Zhang, P. Isola, A. A. Efros, E. Shechtman, and O. Wang, ``The unreasonable effectiveness of deep features as a perceptual metric,'' in \textit{Proc. IEEE/CVF Conf. Computer Vision and Pattern Recognition (CVPR)}, 2018.

\bibitem{heusel2017fid} M. Heusel, H. Ramsauer, T. Unterthiner, B. Nessler, and S. Hochreiter, ``GANs trained by a two time-scale update rule converge to a local Nash equilibrium,'' in \textit{Advances in Neural Information Processing Systems (NeurIPS)}, 2017.

\end{thebibliography}
\end{document}